\documentclass[
]{ceurart}

\begin{document}

\copyrightyear{2020}
\copyrightclause{Copyright for this paper by its authors.\\
  Use permitted under Creative Commons License Attribution 4.0
  International (CC BY 4.0).}

\conference{CHR 2020: Workshop on Computational Humanities Research,
  November 18--20, 2020, Amsterdam, The Netherlands}

\title{Toward a Thermodynamics of Meaning}

\author[1]{Jonathan Scott Enderle}[%
orcid=0000-0003-1901-7921,
]
\ead{enderlej@upenn.edu}
\ead[url]{https://senderle.github.io}
\address[1]{University of Pennsylvania Libraries,
  3420 Walnut St., Philadelphia, PA 19104-6206, United States of America}

\begin{abstract}
  As language models such as GPT-3 become increasingly successful 
  at generating realistic text, questions about what purely text-based 
  modeling can learn about the world have become more urgent. Is text 
  purely syntactic, as skeptics argue? Or does it in fact contain some 
  semantic information that a sufficiently sophisticated language model 
  could use to learn about the world without any additional inputs? This
  paper describes a new model that suggests some qualified answers to 
  those questions. By theorizing the relationship between text and 
  the world it describes as an equilibrium relationship between
  a thermodynamic system and a much larger reservoir, this paper argues
  that even very simple language models do learn structural facts about
  the world, while also proposing relatively precise limits on the nature
  and extent of those facts. This perspective promises not only to answer
  questions about what language models actually learn, but also to explain
  the consistent and surprising success of cooccurrence prediction as a
  meaning-making strategy in AI.
\end{abstract}


\maketitle

\section{Introduction}

Since the introduction of the Transformer architecture in 2017 \cite{vaswani2017}, 
neural language models have developed increasingly realistic text-generation 
abilities, and have demonstrated impressive performance on many downstream 
NLP tasks. Assessed optimistically, these successes suggest that language models, 
as they learn to generate realistic text, also infer meaningful information about 
the world outside of language.

Yet there are reasons to remain skeptical. Because they are so sophisticated,
these models can exploit subtle flaws in the design of language comprehension
tasks that have been overlooked in the past. This may make it difficult to 
realistically assess these models' capacity for true language comprehension. 
Moreover, there is a long tradition of debate among linguists, philosophers, 
and cognitive scientists about whether it is even possible to infer semantics 
from purely syntactic evidence \cite{searle1980}.

This paper proposes a simple language model that directly addresses
these questions by viewing language as a system that interacts with another,
much larger system: a semantic domain that the model knows almost nothing
about. Given a few assumptions about how these two systems relate to one another,
this model implies that some properties of the linguistic system must be shared
with its semantic domain, and that our measurements of those properties
are valid for both systems, even though we have access only to one. But this 
conclusion holds only for some properties. The simplest 
version of this model closely resembles existing word embeddings based on 
low-rank matrix factorization methods, and performs competitively on a
balanced analogy benchmark (BATS \cite{gladkova16}). 

The assumptions and the mathematical formulation of this model are drawn from 
the statistical mechanical theory of equilibrium states. By adopting a materialist
view that treats interpretations as physical phenomena, rather than as abstract 
mental phenomena, this model shows more precisely what we can and cannot infer about meaning
from text alone. Additionally, the mathematical structure of
this model suggests a close relationship between cooccurrence prediction and 
meaning, if we understand meaning as a mapping between fragments of language and 
possible interpretations. There is reason to believe that this line of reasoning 
will apply to any model that operates by predicting cooccurrence, however 
sophisticated. Although the model described here is a pale shadow of a 
hundred-billion-parameter model like GPT-3 \cite{brown20}, the fundamental 
principle of its operation, this paper argues, is the same.

\section{Previous Work}

Most recent work on language modeling builds on the 
word2vec word embedding model \cite{mikolov13b} and its descendants such as GloVe
\cite{pennington14}. These models drew from a longer tradition of distributional 
semantics in linguistics \cite{harris1954} \cite{firth1957} and early machine 
translation research \cite{weaver1955} \cite{masterman1954} \cite{masterman05} 
\cite{gavin18}. The promise of word embedding models for research in the humanities 
was quickly recognized, leading to historical studies of analogical language 
\cite{heuser17} and diachronic lexical change \cite{hamilton16}, but questions 
remained about the utility of embeddings for close humanistic analysis. Word 
embedding models suffer from stability problems, yielding seemingly precise 
answers that change when training input is modified only slightly \cite{antoniak18}, 
and their internal geometric structure is poorly understood \cite{mimno17}. 

Attempts to build a better theoretical understanding of word embeddings have often
focused on exploring the ways different models prove to be mathematically
equivalent in some limit \cite{levy14} \cite{arora16}, or showing the importance 
of preprocessing and hyperparemeter selection. In many cases, with optimal 
hyperparameter choices, factorizing word cooccurrence matrices using SVD and a 
log weighting is sufficient to produce results competitive with state-of-the-art 
models \cite{levy15} \cite{gladkova16}. For these reasons, the claim that word 
embeddings are indeed representations of meaning, and not merely dense 
representations of word cooccurrence, still lacks strong theoretical support. 
On the other hand, even simple coocurrence data seems intuitively to capture 
something about meaning in a way that remains mysterious \cite{arora16}.

More recent language modeling has focused on sequence prediction, either using
recurrent neural networks \cite{peters18} or attention-based mechanisms 
\cite{vaswani2017}. Large language models using the Transformer architecture 
apparently capture rich semantic information usable in a range of downstream 
applications \cite{jawahar19}. But as with word embeddings, there remain empirical 
and theoretical reasons to be skeptical that these models are capturing
information about meaning, rather than performing an extremely sophisticated and 
accurate version of positionally-aware cooccurrence prediction. At least some
attempts to use Transformer models to perform challenging natural language 
comprehension tasks have shown that existing problem datasets contain
subtle linguistic cues that leak information about correct answers \cite{niven19} 
\cite{mccoy19}. These cues have been missed in the past, but with their 
linguistic sophistication, newer models recognize them, producing spurious 
state-of-the-art results without demonstrating true comprehension.

Recent work by Bender and Koller \cite{bender20} provides an even stronger 
theoretical case against the claim that language models infer meaning 
beyond simple cooccurrence. Synthesizing arguments and evidence from 
linguistics and philosophy, including Searle's famous Chinese Room argument 
\cite{searle1980}, Bender and Koller argue that ``the language modeling task, 
because it only uses form as training data, cannot in principle lead to 
learning of meaning.'' Or, in Searle's pithy formulation, the operations of a 
computer have ``syntax but no semantics.'' Bender and Koller's reliance on 
Searle is notable, given that Searle's argument was not against language 
modeling, but against the very possibility of artificial intelligence. Anyone 
who takes his reasoning entirely seriously should forever abandon the notion 
that a computational process could truly comprehend meaning. Yet in their final 
analysis, Bender and Koller back away from Searle's strongest claims, 
acknowledging  that ``if form is  augmented with grounding data of some kind, 
then meaning can conceivably be learned to the extent that the communicative 
intent is represented in that data,'' and that a sufficiently successful 
language model ``has probably learned something about meaning.''
 
\section{Meaning, Cooccurrence, and Thermodynamics}

How can we synthesize these seemingly contradictory bodies of theory and evidence?
It's plausible to claim that language models can never do more 
than predict the way elements of language cooccur in text, since they never see 
any other kind of evidence. And yet even the simplest kinds of 
cooccurrence prediction, such as basic matrix factorization, produce surprisingly
good representations of something that looks intuitively like meaning. Suppose 
that rather than examining the details of particular language models
to see how they differ, and which might be more or less correct, we focus
on what they have in common. Is there some unrecognized connection between
meaning and cooccurrence prediction in all its forms?

This section proposes such a connection based on a model borrowed from
statistical mechanics. Similar models have been applied to machine 
learning problems and algorithms in the past, including language modeling 
tasks \cite{baez12} \cite{stephens10} \cite{srivastava13}. But to the 
author's knowledge, no prior work has used thermodynamic analogies to directly 
investigate the relationship between language and its semantic domain. 

This model begins by treating interpretations as possible configurations of an
unknown physical system. It then constructs a statistical mechanical partition 
function that counts 
the number of interpretations applicable to each fragment of language in a 
corpus. It immediately follows that the Hessian of that function---the matrix of
its mixed second partial derivatives---is a covariance matrix describing word
cooccurrences. The Hessian matrix can be used, in turn, to approximate directional
derivatives of the partition function, which describe the ways the partition
function changes when the meanings of words are slightly modified. These 
directional derivatives are word vectors, with all the expected properties.

\subsection{Model Assumptions}

Setting up our model requires some odd assumptions about how language 
works. To begin with, it requires that we assume that meaning is quantifiable
in the most naive way. It's not uncommon in colloquial speech to talk about the 
amount of meaning a phrase has, without specifying what the phrase means. Some 
phrases, we might say, are meaningless; others are full of meaning. To construct
a statistical mechanical model of meaning, it is useful to assume that this is a 
perfectly correct way of quantifying meaning, and that, so quantified, meaning 
is a conserved value that plays the same role as energy in a typical 
thermodynamic ensemble.

As long as we are making extravagant assumptions, let's also assume that for a 
given linguistic system and an associated semantic domain, words have a stable 
average capacity for holding meaning, and that word counts are conserved values 
just like energy, so that a combined linguistic system and associated semantic
domain contains an unknown but fixed number of copies of every possible word. 
Leaving aside the linguistic significance of these assumptions for a moment, we 
can skip ahead by recognizing them as formally equivalent to the assumptions
made in the construction of the grand canonical ensemble.

\subsection{The Grand Canonical Ensemble}

In classical thermodynamics, the grand canonical ensemble describes a system of 
particles---such as a container of gas---that is in thermodynamic and chemical
equilibrium with a much larger system, a reservoir of energy and particles. 
Concretely, this means that the temperature of the gas in the container is the 
same as its surroundings (assumed to be homogenous, and far larger than the 
container), and that the container can exchange particles with its surroundings, 
but at a steady state, so that it is as likely to lose a particle as to gain a 
particle at any given moment. Furthermore, both the amount of energy and the 
number of particles shared between the container and its surroundings are 
fixed---energy and particle number are conserved values.

To understand the behavior of this ensemble, we begin by imagining that we could
track the exact position and momentum of every particle in the system (container), 
as well as the exact position and momentum of every particle in the reservoir 
(surroundings of the container). At a given instant in time, these values 
constitute a ``microstate.'' Since we have both a system and a reservoir, we can 
divide a single microstate into parts, considering just the microstate of the 
system, or just the microstate of the reservoir. We can also determine that 
certain system microstates are incompatible with certain reservoir microstates, 
because the combination would violate a conservation law. In other words, for
some pairs of system microstate and reservoir microstate to coexist, energy or
particles would have to be created or destroyed, which would violate the rule 
that energy and particle number are conserved values.

If we rule out all system-reservoir microstate pairs that are not compatible 
($\not\leftrightarrow$), and assume that all reservoir microstates are equally
likely---an acceptable approximation when the reservoir is far larger than the 
system---then we can approximate the probability of a given system microstate 
$s_i$ by counting the number of reservoir microstates that are compatible 
($\leftrightarrow$) with it. Using Iverson brackets ($[i = j] = \delta_{ij}$) 
we can say

\begin{equation}
p_i \propto \sum\limits_{j} [r_j \leftrightarrow s_i]
\end{equation}

To recover the probability itself, we can divide by the sum over all $s$:

\begin{equation} \label{eq:countprob}
p_i = \frac{\sum\limits_{j} [r_j \leftrightarrow s_i]}
              {\sum\limits_{j,k} [r_j \leftrightarrow s_k]}
\end{equation}

These sums are very large, and it's not clear how to calculate them. But it turns
out we don't need to. Given a few standard assumptions from thermodynamics, our
assumptions about conserved quantities, and a bit of calculus, it's possible to use 
them to derive the following function:

\begin{equation}
Z = \sum\limits_{i} e^{\beta(\mu N_i - E_i)}
\end{equation}

This is the grand canonical partition function. From it, we can then directly
calculate the probability of system microstate $i$ like so:

\begin{equation}
p_i = \frac{e^{\beta(\mu N_i - E_i)}}{Z}
\end{equation}

This formula tells us, first, that at a given fixed temperature $T$ determined by 
$\beta = \tfrac{1}{k_BT}$, system microstates with more energy ($E_i$) are
less probable, because they are compatible with fewer reservoir microstates.
It also tells us that for any given energy level, system microstates containing 
more particles ($N_i$) with a higher chemical potential ($\mu$) are more probable.
This is because given two systems with the same energy, the system with a higher
overall potential has a higher energy capacity.

This partition function can be extended to systems that have multiple kinds
(``species'') of particles. In that case, each species has its own chemical
potential and count. For a system with $k$ different species

\begin{equation} \label{eq:multpart}
Z = \sum\limits_{i} 
   e^{\beta(\mu_1 N_{1,i} + \mu_2 N_{2,i} + ... + \mu_k N_{k,i} - E_i)}
\end{equation}
\begin{equation} \label{eq:multiprob}
p_i = \frac{e^{\beta (\mu_1 N_{1,i} + \mu_2 N_{2,i} + ... + \mu_k N_{k,i} - E_i)}}
           {Z}
\end{equation}

\subsection{From Compatibility to Interpretation}

What does all this have to do with language? The first hint that the grand
canonical partition function might have some usefulness as a model for language
is that energy and meaning (in the naive quantitative sense described above) both
impose similar compatibility constraints on the system and reservoir. Just as 
higher-energy states in the system correspond to fewer possible reservoir states,
more meaningful sentences correspond to fewer possible interpretations. A
statement with less meaning has less precision, while a statement with more 
meaning has more precision, eliminating a larger number of possible 
interpretations. The line of reasoning is similar for particle species and words. 
Just as a particle species with higher chemical potential has higher energy 
capacity, a word with a higher ``semantic potential'' has a higher capacity for 
meaning.

Consider, for example, the sentence ``It stinks.'' Then compare it to ``On January
15, 2008, a rainfall of 110mm was recorded in the city of Dubai.'' The specific
interpretations these sentences can be given will depend on context, and in some
contexts, ``It stinks'' might be a meaningful and precise sentence. But 
on balance, we should expect ``It stinks'' to be less meaningful than 
``On January 15...,'' both because it contains fewer words, and because the words
it contains are less precise than words like ``rainfall'' and ``Dubai.''

Although this is a simple way of thinking about meaning, it is not as simplistic
as it may seem at first. Consider the sentence ``Ask for me tomorrow, and you 
shall find me a grave man,'' as uttered by a dying Mercutio. One might think that 
by the logic above, this sentence would be made less meaningful by the presence 
of an ambiguous word, ``grave,'' here meaning either ``serious'' or ``a place 
of burial.'' But a more careful analysis leads to a different conclusion. If 
these two senses were available independently, and the sentence could be 
properly interpreted in two different ways, it would indeed be less meaningful 
because of this ambiguity. In this context, however, choosing just one of those 
senses to the exclusion of the other would yield a misreading of the sentence. 
It does not invite two different possible interpretations; it invites one 
interpretation that combines together two distinct concepts both conveyed by 
the word ``grave.'' By eliminating interpretations that do not combine these 
two senses together, this sentence uses ambiguity to achieve a higher degree 
of precision. Analyzed this way, literary language is often likely to be more 
precise and meaningful than everyday language, despite sometimes having greater 
surface ambiguity.

If we translate these ideas into a mathematical form, and start thinking 
about compatibility ($\leftrightarrow$) as a semantic relationship, then
equation \ref{eq:countprob} says roughly that the probability of a given 
sentence ($s_i$) is equal to the number of interpretations ($r_{j,..,k}$) it has, 
divided by the number of interpretations that all possible grammatically correct 
sentences have. The refinement of that equation to equation \ref{eq:multiprob} 
now says that sentences with more meaning are less probable, because they are
compatible with fewer interpretations, and that for any given degree of 
meaningfulness, sentences with a higher semantic potential are more probable. 
(That is, precise sentences are harder to write, but it's easier to write a precise 
sentence with more words, and it's harder to pack all your meaning into just a few 
very precise words.)

\subsection{From Ensembles to Vectors}

Most word embedding models generate word vectors by using a supervised or
semi-supervised model to predict cooccurrences, and the vectors themselves aren't
significant outside that predictive frame. But the picture is quite different
for statistical-mechanical models such as this one. One of the most elegant
properties of partition functions is that a wide range of thermodynamic 
quantities can be expressed directly as partial derivatives of the partition
function or its logarithm.

For example, suppose we would like to determine the number of particles of a 
particular kind present in all possible states of our system ($N_k$), and take the 
average. We can calculate that value by taking the partial derivative of the 
logarithm of equation \ref{eq:multiprob} with respect to the chemical potential 
of that species, and dividing out $\beta = \tfrac{1}{k_BT}$.

\begin{equation}
\langle N_k \rangle = 
    \frac{1}{\beta}
    \frac{\partial \ln}{\partial \mu_k} Z(\mu_k)
\end{equation}

Since $\tfrac{\partial \ln}{\partial x} f(x) = 
       \tfrac{\partial}{\partial x} f(x) / f(x)$, this simplifies to a 
probability-weighted sum of $N_k$ counts divided by $\beta$, effectively an 
arbitrary constant multiplier. Shifting it to the left hand side of the equation
gives

\begin{equation}
\beta \langle N_k \rangle
    = \frac{\frac{\partial}{\partial \mu_k} Z(\mu_k)}{Z(\mu_k)}
    = \sum\limits_{i} N_{k,i}
    \frac{e^{\beta(\mu_1 N_{1,i} + \mu_2 N_{2,i} + ... 
                             + \mu_k N_{k,i} - E_i)}}
         {Z}
    = \sum\limits_i N_{k,i} p_i
\end{equation}

This line of reasoning can be extended to second partial derivatives. The variance
of $N_k$ is given by

\begin{equation}
    \beta \left[\langle N_k^2 \rangle - 
          \langle N_k \rangle^2\right] = 
    \frac{\partial^2 \ln}{\partial \mu_k^2} Z(\mu_k)
\end{equation}

Similarly, the covariance of $N_k$ and $N_j$ is a mixed partial derivative.

\begin{equation}
    \beta 
    \left[\langle N_k N_j \rangle - 
          \langle N_k \rangle \langle N_j \rangle\right] = 
    \frac{\partial^2 \ln}{\partial \mu_k \mu_j} Z(\mu_k, \mu_j)
\end{equation}

These last two equations can be used to construct a matrix that has two simultaneous
meanings. It is, first, a covariance matrix that describes the way particle counts 
are correlated with one another in the system. But it is also a Hessian matrix of 
second partial derivatives, meaning that it describes the way small modifications
to the chemical potential terms change the overall partition function, shifting its
energy balance across all possible system microstates. This means that even if we
can't construct the partition function itself, we can in principle measure the 
covariance of particles empirically, and use the resulting matrix to reconstruct 
information about the partition function and the thermodynamic ensemble it describes.

If we translate this into linguistic terms, we find that by taking empirical
measurements of word cooccurrence, we are also constructing the Hessian of a
linguistic partition function that describes how changes to the meaning of one
word affect the meaning of another. The columns of that matrix are word vectors.
When two columns are similar, small modifications to the meanings of the associated 
words have similar effects on the language as a whole. That is what it means, 
in the context of this model, for two words to be similar. Line integrals through 
the Hessian field in a given neighborhood can also be approximated by adding and 
subtracting these vectors, giving a more precise interpretation to the formulas
used to represent analogies. Analogies are valid when they correspond to two 
different line integrals through a conservative Hessian tensor field, beginning
at the same point and ending close to the same point, and therefore having
similar final values.

\subsection{Implementation}

Constructing a practical implementation of this model requires that we determine 
the values for two sets of parameters: the potential for each word in the vocabulary,
and the energy level for each sentence. The simplest approach to this problem is
to set all potential terms to zero, and all energy terms to one. The covariance
matrix that results from these choices is identical to the one given by directly
counting word cooccurrences. Alternative schemes will change the weights given
to each of the sentences, yielding a modified covariance matrix that is likely
to give better meaning representations. For performance reasons, some form of
dimension reduction is also necessary, but has no theoretical significance at 
all. In practice, random projection (as in \cite{schmidt18}) works well, 
especially after implementing some of the preprocessing and hyperparameter 
selection recommendations in \cite{levy15}, which may be compensating for the 
deficiencies that result from setting the energy and potential terms to constants.

The problem of selecting energy and potential terms in a more principled way is
left to other work. But it is worth considering briefly, since it illustrates
some interesting properties of the model. First, in this model, the same sequence of words could 
appear twice with different energy levels, and therefore 
different probabilities, depending on context. Second, there may be 
a way to make predictions that link the semantic potential of given terms to known
lexicographical properties of those terms, such as the approximate number of
senses the word has. And finally, the partition function described here is
not the only possible partition function that might be applied to language. 
Partition functions based on word pairs, sequences, or even attention mechanisms 
could be used to model language within this framework, all broadly interpretable 
in the same way.

\section{Discussion}

Few of the ideas presented here are new. The fact that word vectors contain
distributional information that allows them to measure word similarity has
been known for decades. Ideas from statistical mechanics have been applied to 
language modeling, machine learning, and information retrieval problems for
decades. And for the last few years, there has been a steady stream of work
demonstrating that language model X reduces to language model Y in some limit. 
But none of this has shown how these models could capture information
about interpretation or meaning. Trained on linguistic form alone, these models 
have no evidence showing how linguistic forms map to mental models, 
concepts, narratives, or any other representations of things outside of language.

The claim that statistical models can infer things about meaning from linguistic
form alone thus faces a high burden of proof. And while there has been a 
proliferation of models that do appear to support that claim, they all work
on slightly different principles, and produce slightly different results. This 
undercuts attempts at meta-induction; many small bodies of evidence based on
different principles of operation do not add up to one large body of evidence.
And so justified skepticism remains.

What is new about the model proposed here is that it is general enough to 
explain the success of many of these models without reference to the details
of their operation. Fundamentally, any model that is able to predict linguistic
cooccurrences can be reinterpreted as an implicit partition function along the
lines proposed here. So reinterpreted, we can argue that distributional information
about language is linked by a precise mathematical structure to specific facts
about how words signify. Those facts are limited; they do not include any 
information about \textit{what} words, sentences, or longer fragments of language 
talk about. But they do include information about \textit{how many} 
interpretations might be applied to those units of language, and how those 
interpretations correlate with one another at a macroscopic level.

What unites all of these models, under this theory, is that they effectively
assume that meaning, quantified appropriately, is conserved, and that units of 
language---be they letters, words, n-grams, or longer phrases---are also
conserved. It's not yet clear what these assumptions might mean in linguistic
terms, but they are crucial to the derivation of a partition function that can
relate the statistics of linguistic form to an unknown reservoir of meaning.

These models must also make a third assumption: language exists in a state of 
equilibrium with its reservoir of meaning. That assumption is unlikely to hold
in general. If this way of thinking about language modeling is sound, then
an important project will be to understand when the assumption of equilibrium
is justified, and when it is not. It's likely that during periods of rapid
linguistic change, for example, the equilibrium assumption will not be valid. 
In that case, methods that can model far-from-equilibrium systems will be
required. Since non-equilibrium thermodynamics is a field still in its infancy
\cite{england15}, there will be much work to be done, and many tasks that remain
impossible without domain expertise. Nonetheless, a deeper understanding of the 
meaning of these assumptions promises to clarify when and how language models
can infer meaning from linguistic form alone. 

\bibliography{bibliography}

\end{document}